\documentclass{article}
\usepackage{iclr2027_conference,times}
\iclrfinalcopy
\usepackage[T1]{fontenc}
\usepackage[utf8]{inputenc}
\usepackage{amsmath,amssymb,amsthm,booktabs,array}
\usepackage{algorithm,algpseudocode}
\newtheorem{proposition}{Proposition}
\usepackage{graphicx,placeins}
\graphicspath{{./}{figures/}}
\usepackage{tikz}
\usetikzlibrary{arrows.meta,positioning,calc}
\usepackage{xurl}
\usepackage[hidelinks]{hyperref}
\title{Feedback-Calibrated Protein Optimization with Batch-Aligned Tail Arbitration}
\author{\begin{minipage}{0.96\textwidth}\centering
\textbf{Zefeng Lin}$^{1}$ \quad \textbf{Xianyong Fang}$^{3}$ \quad \textbf{Tianfan Fu}$^{2}$ \quad \textbf{Xiaohua Xu}$^{1}$\\[0.6ex]
{\small $^{1}$School of Computer Science and Technology, University of Science and Technology of China, Hefei, Anhui, China\\
$^{2}$State Key Laboratory for Novel Software Technology at Nanjing University, School of Computer Science, Nanjing University, Nanjing, Jiangsu, China\\
$^{3}$School of Computer Science and Technology, Anhui University, Hefei, Anhui, China}
\end{minipage}}
\hypersetup{pdftitle={Feedback-Calibrated Protein Optimization with Batch-Aligned Tail Arbitration},pdfauthor={Zefeng Lin, Xianyong Fang, Tianfan Fu, Xiaohua Xu}}
\begin{document}
\maketitle
\fancyhead[L]{Preprint}
\begin{abstract}
Protein optimization aims to discover high-fitness sequences under a limited experimental budget. Existing machine-learning methods use task-specific predictors, biological priors, or ranking-aware objectives to guide which variants are tested in the next experimental round. 
However, these methods cannot adapt to shifts in the reliability of predictive evidence as measurements accumulate and ensure the correct ranking of key high-fitness candidates.
To address these challenges, we propose \textbf{Batch-Aligned Tail Arbitration (BATA)}, which uses experimental feedback to adaptively combine prior-informed and task-specific rankings for next-batch selection, with calibration focused on the batch-aligned high-fitness region. Across measured GB1, PABP, and TrpB landscapes, BATA achieves the best mean task rank (1.67) in final best fitness after 480 measurements. Controlled comparisons further show task-dependent gains from high-fitness calibration and batch alignment. 
Our work introduces feedback-calibrated predictor arbitration, where experimental feedback dynamically determines how predictive evidence guides next-batch selection, opening a new direction for protein optimization.
\end{abstract}
\section{Introduction}
Protein optimization aims to discover amino acid sequences with improved functional properties for applications such as therapeutics, industrial biotechnology, and environmental engineering \citep{notin2024functional,listov2024opportunities}. The performance of a protein for a target function is commonly measured by its \emph{fitness}, and the relationship between protein sequences and their fitness values defines a \emph{protein fitness landscape} \citep{freschlin2022navigate}. However, protein fitness landscapes are vast and often rugged, while the fitness of a candidate sequence can usually be determined only through costly experimental measurements \citep{yang2024opportunities,sandhu2025fitness}. To reduce experimental costs under a limited assay budget, modern protein optimization commonly uses machine-learning models to predict fitness from available measurements and iteratively select promising variants for subsequent experiments \citep{rapp2024selfdriving,herrera2025bestpractices}.
Existing machine-learning methods for protein optimization can be broadly grouped into three categories: (1) Active-learning and Bayesian-optimization methods iteratively train a fitness predictor on measured variants and use its predictions to select candidates for the next round of experiments \citep{yang2025active,bal2025gameopt}. (2) Prior-guided methods incorporate evolutionary or generative sequence priors to guide the search when task-specific measurements are limited \citep{amin2025clonebo}. (3) Ranking-aware methods focus more directly on ordering promising variants so that high-fitness candidates can be prioritized for experimental evaluation \citep{karimi2025distillation,xu2026rank}. Despite their different designs, these approaches all rely on predictive evidence to guide which variants should be tested in the next experimental round.

\begin{figure}[t]
\centering
\includegraphics[width=\linewidth]{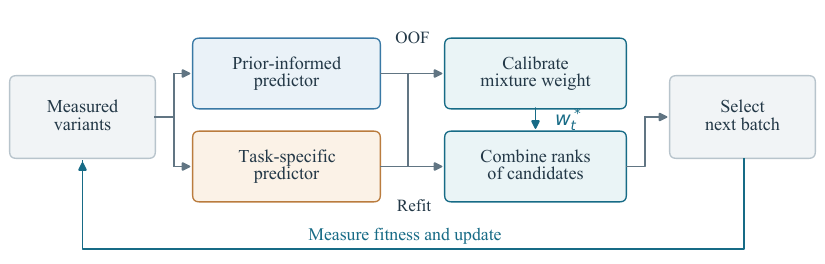}
\caption{\textbf{BATA overview.} Measured fitness trains both predictors and calibrates their relative influence. Five-fold out-of-fold (OOF) predictions are converted to ranks to fit a mixture weight on the observed high-fitness region. After refitting on all measurements, the predictors rank unmeasured candidates. Their weighted ranks guide batch selection, and the new measurements start the next round.}
\label{fig:bata_overview}
\end{figure}

However, the predictive evidence used by these methods is not equally reliable throughout an optimization: (1) When only a few experimental measurements are available, a task-specific predictor is trained on limited data and may be unreliable, especially on sparse or difficult fitness landscapes \citep{ghaffari2024robust,ssmula2025,ibarraran2026efficient}. In this stage, broader biological priors can provide useful guidance. (2) As more measurements are collected, the task-specific predictor can become more informative, while the relative contribution of the biological prior may change across optimization rounds. (3) Even accurate prediction across all measured variants does not necessarily guarantee correct ranking among the key high-fitness candidates considered for the next batch, which will decide the success of the next experiment. Therefore, the main challenge is to use experimental feedback to determine how much each source of predictive evidence should influence next-batch selection, while correctly ranking the most promising high-fitness candidates.
To address these challenges, we propose \textbf{Batch-Aligned Tail Arbitration (BATA)}, a method for feedback-calibrated protein optimization (Figure~\ref{fig:bata_overview}). 
For the first challenge, BATA retains both a prior-informed and task-specific predictor, allowing the biological prior to provide complementary guidance when task-specific measurements are still limited. Their predictions are converted to percentile ranks so that the two sources of evidence can be compared on a common scale. 
For the second challenge, BATA uses five-fold out-of-fold (OOF) predictions on measured variants to assess the current usefulness of each predictor. It then learns a mixture weight from the measured fitness values and updates this weight after every batch, allowing the relative contribution of the two predictors to change as experimental evidence accumulates. 
For the third challenge, BATA calibrates this mixture weight specifically on the highest-fitness measured variants instead of all observations. The calibration cutoff is set to the next batch size, and a logarithmically discounted cumulative gain (DCG) weight places greater emphasis on variants near the top of the fitness ranking. The obtained weighted rank objective has a closed-form solution, and the learned mixture is used to rank unmeasured sequences for the next experimental batch.

Our main contributions are summarized as follows: 

(1) We formulate feedback-calibrated protein optimization as a decision problem in which experimental measurements determine not only how task-specific predictors are updated, but also how prior-informed and task-specific evidence should contribute to next-batch selection. 
(2) We propose BATA, which evaluates both predictors using out-of-fold evidence, adaptively learns their mixture weight from measured fitness, and calibrates this weight on a batch-aligned high-fitness region using a DCG-weighted rank objective with a closed-form solution, to adjust the relative contribution of different evidence while prioritizing the promising high-fitness candidates.

(3) Across GB1, PABP, and TrpB, BATA achieves the best mean task rank (1.67) in final best fitness over 35 matched initializations per task, while controlled comparisons reveal when high-fitness calibration, batch alignment, and prior-informed evidence are useful. 
Our work extends the role of experimental feedback from updating predictors to arbitrating complementary predictive evidence, offering a new direction for feedback-calibrated protein optimization.

\section{Related Work}
\paragraph{Learning from experimental feedback.}
Machine-learning-assisted directed evolution uses measured protein variants to fit a fitness predictor and select candidates for the next experiment \citep{wu2019machine,yang2019machine}. EVOLVEpro combines protein-language-model representations with task-specific regression, while ProSpero updates a fitness predictor from feedback to guide search beyond wild-type neighborhoods \citep{jiang2025evolvepro,prospero2025}. EvoMax further combines complementary models for protein evolution with sparse experimental data \citep{wan2026evomax}. These studies show how experimental feedback can improve predictions and guide search. 
BATA extends this feedback loop by using measured outcomes not only to update predictors but also to recalibrate how predictive evidence guides the next experimental batch.
\paragraph{Searching protein fitness landscapes.}
Protein optimization also benefits from methods that change the search space or the way it is explored. GGS smooths a learned fitness landscape, and LatProtRL uses reinforcement learning to search a continuous protein representation \citep{kirjner2024ggs,lee2024latprotrl}. KnowRLM incorporates biological knowledge into reinforced language modeling, while VLGPO combines a generative prior with fitness guidance in a latent-space optimization method \citep{wang2024knowrlm,bogensperger2025vlgpo}. 
In contrast, BATA leaves the representation and search space unchanged and instead focuses on how predictive evidence is combined within batch selection.
\paragraph{Priors and sequential decisions.}
Prior-guided optimization studies how existing knowledge can support a sequence of experimental decisions. In Bayesian optimization, DynMeanBO changes the influence of an expert prior, while ToSFiT and BOLT use language models for sequential optimization and transfer across tasks \citep{qu2026dynmeanbo,menet2026tosfit,zeng2026bolt}. For proteins, Data Distillation and REAP learn candidate preferences or rankings to prioritize promising variants \citep{karimi2025distillation,xu2026rank}.
Building on these, BATA uses a prior-informed predictor to provide biological guidance when task-specific measurements are still limited.

\section{Feedback-Calibrated Protein Optimization}
\label{sec:feedback_calibrated}
Let $\mathcal X=\{x_1,\ldots,x_N\}$ be a finite protein sequence candidate pool and $f(x)$ its experimental fitness. At round $t$, the available data are
\begin{equation}
\mathcal D_t=\{(x_i,y_i):x_i\in\mathcal M_t,\ y_i=f(x_i)\},
\end{equation}
where $\mathcal M_t$ contains the sequences measured so far. The optimizer chooses $B$ distinct sequences $\mathcal B_{t+1}\subseteq\mathcal X\setminus\mathcal M_t$ for the next batch. Their fitness values are revealed together and added to form $\mathcal D_{t+1}$. Unmeasured fitness values are unavailable for fitting, calibration, and selection.
The goal of the protein optimization task is to maximize the best observed fitness under a fixed measurement budget. We start with 96 measurements and add four batches of $B=96$. The primary endpoint, Final@480, is the highest fitness found among these 480 measurements:
\begin{equation}
\mathrm{Final}@480=\max_{x\in\mathcal M_4} f(x).
\end{equation}
The secondary metric, Query-AUC, is the normalized area under the best-so-far fitness curve from 96 to 480 measurements (Appendix~\ref{app:reproducibility}). A higher Query-AUC indicates earlier discovery of high-fitness variants.
The prior-informed predictor and the task-specific predictor learn from the same revealed fitness measurements, but use different sequence representations and prediction models. Their rankings can therefore disagree. We define \emph{feedback calibration} as using measured data to estimate how these rankings should be combined. Only $B$ candidates can be tested in the next batch, making their position near the top of the ranking especially relevant. Therefore, calibration should account for both the observed high-fitness region and the next batch size.

\section{Batch-Aligned Tail Arbitration}
\label{sec:bata}
At each round, BATA uses measured fitness to set the balance between the prior-informed predictor and the task-specific predictor. It chooses a mixture weight that brings their combined ranking closest to the observed ranking of high-fitness variants. The same weight combines their rankings over unmeasured sequences for next-batch selection (Figure~\ref{fig:bata_overview}).

\subsection{Prior-informed and task-specific predictors}
\label{sec:bata_views}\label{sec:bata_oof}
The prior-informed predictor is an ensemble of Ridge regression models trained on direct coupling analysis (DCA) features, which summarize evolutionary sequence statistics \citep{morcos2011dca}. The task-specific predictor uses the sequence representation and prediction model specified for each benchmark in Appendix~\ref{app:reproducibility}. Both predictors are fitted to the same measured fitness data. As measurements accumulate, the prior-informed predictor is refitted while retaining the evolutionary information in its DCA features.
At round $t$, reindex the $n_t=|\mathcal D_t|$ measured variants by $i=1,\ldots,n_t$ and collect their fitness values in $\mathbf y=(y_i)_{i=1}^{n_t}$. To obtain out-of-fold (OOF) predictions, we divide these variants into five folds. For each fold, the prior-informed and task-specific predictors are trained on the other four folds and predict the held-out variants. Averaging the ensemble members gives two OOF vectors, $\hat{\mathbf y}^{\rm prior}$ and $\hat{\mathbf y}^{\rm task}$, with one prediction per variant. Each prediction is made without training on that variant's fitness value. We convert the fitness and OOF vectors to percentile ranks:
\begin{equation}
 \mathbf r^y=\rho(\mathbf y),\qquad \mathbf r^p=\rho(\hat{\mathbf y}^{\rm prior}),\qquad
 \mathbf r^a=\rho(\hat{\mathbf y}^{\rm task}).
 \label{eq:percentile_ranks}
\end{equation}
For a real vector of length $n\ge1$, $\rho:\mathbb R^n\to[1/n,1]^n$ assigns average ascending ranks and divides them by $n$. Thus, higher values receive higher percentile ranks. The components $r_i^y$, $r_i^p$, and $r_i^a$ denote the fitness, prior-informed, and task-specific ranks of measured variant $i$.

\subsection{Calibration on the observed high-fitness region}
\label{sec:bata_tail}
To match calibration to the next batch size, set the fitness-rank cutoff to $k_t=\min(B,n_t)$. We assign each variant its average descending fitness rank $p_i$. We weight the top-ranked variants using the logarithmic discount from discounted cumulative gain (DCG), which gives more weight to items near the top of a ranking \citep{jarvelin2002dcg}:
\begin{equation}
 q_i=\begin{cases}
 1/\log_2(p_i+1),&p_i\le k_t,\\
 0,&p_i>k_t.
 \end{cases}
 \label{eq:tail_weights}
\end{equation}
Higher-fitness variants receive larger weights, and variants below the cutoff receive zero weight. Average ranks treat tied fitness values equally, so a tie at the cutoff can make the number of nonzero weights differ from $k_t$. All 96 initial measurements are within the first-round cutoff. Later rounds focus on the high-fitness region of the growing measured set.
Let $w\in[0,1]$ be the weight of the prior-informed predictor; the task-specific predictor has weight $1-w$. Their combined rank score is $r_i(w)=wr_i^p+(1-w)r_i^a$. We choose $w$ to minimize the squared error between the combined rank and the observed fitness rank, using the DCG discounts $q_i$ as sample weights:
\begin{equation}
 \mathcal L_{\rm BATA}(w)=\sum_{i=1}^{n_t}q_i\bigl[r_i^y-wr_i^p-(1-w)r_i^a\bigr]^2.
 \label{eq:bata_objective}
\end{equation}

\subsection{Closed-form weight and candidate selection}
\label{sec:bata_select}
To compute the mixture weight, set $d_i=r_i^p-r_i^a$ and $z_i=r_i^y-r_i^a$. The loss is a one-dimensional convex quadratic with constrained solution
\begin{equation}
 w_t^*=\Pi_{[0,1]}\!\left(\frac{\sum_iq_id_iz_i}{\sum_iq_id_i^2}\right),
 \label{eq:bata_weight}
\end{equation}
Here $\Pi_{[0,1]}$ clips a value to $[0,1]$. If the denominator is zero, we set $w_t^*=0$. The implementation applies this rule when the denominator is no greater than $\epsilon_{\rm mach}$, the float64 machine epsilon. Larger values of $w_t^*$ give the prior-informed predictor more influence; smaller values give the task-specific predictor more influence. BATA recomputes $w_t^*$ after each batch, so new fitness measurements can change the balance between the predictors.
After estimating $w_t^*$, we refit both predictors on all measured variants. Let $g^{\rm prior}_{t,m}$ and $g^{\rm task}_{t,m}$ be their prediction functions for member $m=1,\ldots,L$, each mapping a sequence in $\mathcal X$ to a real-valued fitness prediction. A single-model predictor is reused across all $L$ members of the other ensemble. We reindex the remaining pool $\mathcal U_t=\mathcal X\setminus\mathcal M_t$ by $x_1,\ldots,x_{N-n_t}$ and form prediction vectors $\mathbf g^{\rm prior}_{t,m}=(g^{\rm prior}_{t,m}(x_j))_{j=1}^{N-n_t}$ and $\mathbf g^{\rm task}_{t,m}$ in the same way. We rank each member's vector over this pool and combine its components:
\begin{equation}
 s_{j,m}=w_t^*\bigl[\rho(\mathbf g^{\rm prior}_{t,m})\bigr]_j
 +(1-w_t^*)\bigl[\rho(\mathbf g^{\rm task}_{t,m})\bigr]_j.
 \label{eq:fused_pool}
\end{equation}
These pool ranks stay fixed while the batch is selected. GB1 and TrpB use Thompson-style selection \citep{russo2018tutorial}: an ensemble member is sampled for each candidate choice, and its highest-scoring remaining sequence is selected. PABP greedily selects sequences using the mean combined score. The selection rule returns $B$ distinct sequences before any fitness values from that batch are measured. In our experiments, $T=4$ such rounds follow the initial measurements.

\begin{algorithm}[t]
\caption{BATA for sequential protein optimization}
\label{alg:bata}
\begin{algorithmic}[1]
\Require Candidate pool $\mathcal X$, initial data $\mathcal D_0$, batch size $B$, rounds $T$, selection rule $\mathcal A$
\For{$t=0,\ldots,T-1$}
 \State Let $\mathcal M_t$ be the sequences in $\mathcal D_t$.
 \State Obtain five-fold OOF predictions from both predictors on $\mathcal D_t$.
 \State Compute percentile ranks $r^y,r^p,r^a$ and weights $q$ (Eqs.~\ref{eq:percentile_ranks}--\ref{eq:tail_weights}).
 \State Compute $w_t^*$ by Eq.~\ref{eq:bata_weight}, with the stated numerical convention.
 \State Refit both predictors on $\mathcal D_t$ and compute pool scores $s$ by Eq.~\ref{eq:fused_pool}.
 \State Select $\mathcal B_{t+1}=\mathcal A(s,B)\subseteq\mathcal X\setminus\mathcal M_t$.
 \State Reveal the full batch and add its measurements to form $\mathcal D_{t+1}$.
\EndFor
\State \Return The highest-fitness measured variant.
\end{algorithmic}
\end{algorithm}

\subsection{Properties of the calibration rule}
\label{sec:bata_properties}
\begin{proposition}[Optimality for the observed rank loss]
\label{prop:bata_optimality}
For fixed ranks and nonnegative $q_i$, Eq.~\ref{eq:bata_weight} uniquely minimizes Eq.~\ref{eq:bata_objective} over $[0,1]$ when $\sum_iq_id_i^2>0$. If this sum is zero, the loss is constant and $w_t^*=0$ is a minimizer.
\end{proposition}
\begin{proposition}[Score-scale invariance]
\label{prop:bata_invariance}
With a fixed tie convention, strictly increasing transformations of either predictor's fixed, ensemble-averaged OOF vector or each member's fixed pool vector leave the calibration weight and fused percentile-rank scores unchanged.
\end{proposition}
Therefore, BATA can calibrate predictors with different score units by solving one convex problem per round. Appendix~\ref{app:calibration_properties} provides proofs of both properties.

\section{Experiments}
\label{sec:experiments}

\subsection{Experimental setup}
\label{sec:setup}
We evaluate BATA by replaying sequential optimization on measured GB1, PABP, and TrpB fitness landscapes \citep{melamed2013pabp,wu2016adaptation,johnston2024combinatorially}. Each campaign is one optimization run: 96 initial measurements followed by four batches of 96, for 480 unique measurements. We report the Final@480 and Query-AUC metrics defined in Section~\ref{sec:feedback_calibrated}. The main cohort, Main-35, uses 35 matched initialization sets per task. A separate sensitivity cohort uses 70 initializations for GB1 and TrpB and retains the 35 PABP initializations. Model fitting and candidate selection use only fitness values revealed by the current round.
We compare BATA with ALDE, EVOLVEpro-650M, REAP100-650M, and a random forest (RF). These baselines use active learning, protein-language-model regression, rank-guided learning, and standard supervised selection, respectively \citep{jiang2025evolvepro,yang2025active,xu2026rank}. All methods within a task and cohort share the initialization identities and the 480-measurement budget. BATA's prior-informed predictor uses a DCA-based Ridge ensemble. Its task-specific predictor uses one-hot Ridge models on GB1 and ESM2-650M random forests on PABP and TrpB. Appendix~\ref{app:reproducibility} lists the ensemble sizes, selection rules, and analysis cohorts. Tables give means and sample standard deviations. Curve bands give pointwise 95\% bootstrap confidence intervals (CIs), obtained by resampling complete campaigns.

\subsection{Optimization quality and discovery speed}
\label{sec:overall}
\begin{figure}[t]
\centering
\includegraphics[width=\linewidth]{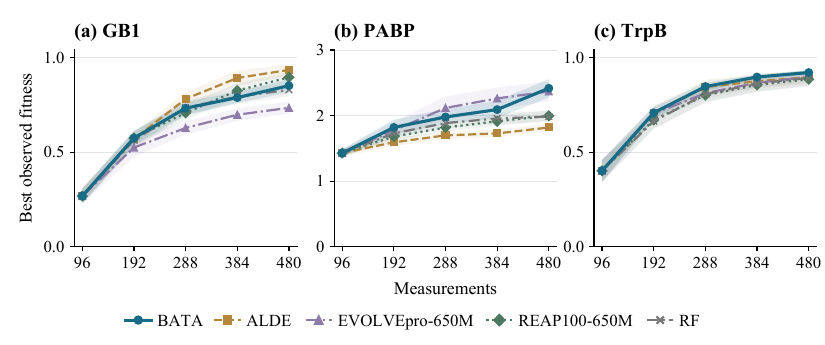}
\caption{\textbf{Optimization trajectories.} Lines show mean best-observed fitness and bands show pointwise 95\% bootstrap confidence intervals over 35 matched campaigns per method and task (Main-35). Points mark the initial measurements and each completed batch. Fitness values use each task's native scale.}
\label{fig:main_performance}
\end{figure}

BATA has the lowest mean Final@480 rank across the three Main-35 tasks (1.67; Table~\ref{tab:main_performance}). It achieves the highest mean Final@480 on PABP and TrpB, while ALDE leads on GB1. BATA also has the lowest mean Query-AUC rank (2.00). Figure~\ref{fig:main_performance} tracks best-observed fitness over the five measurement milestones. On PABP, EVOLVEpro discovers high-fitness variants earlier, as measured by Query-AUC, but BATA reaches a higher Final@480. Appendix~\ref{app:additional_results} reports the full Query-AUC results and per-campaign distributions.
\begin{table}[t]
\centering\small
\caption{\textbf{Main optimization results.} Final@480 is the best observed fitness after 480 measurements. Entries give means $\pm$ sample standard deviations over 35 matched initializations per task (Main-35). The last two columns average the task-wise ranks of Final@480 and Query-AUC, the area-under-curve metric for discovery speed. Lower mean ranks are better. Bold marks the best task mean or mean rank.}
\label{tab:main_performance}\label{tab:main35_full}
\setlength{\tabcolsep}{3pt}
\begin{tabular}{lccccc}
\toprule
Method & GB1 & PABP & TrpB & Final rank & AUC rank\\
\midrule
BATA & $0.8512\pm0.1083$ & $\mathbf{2.4169\pm0.4232}$ & $\mathbf{0.9209\pm0.0422}$ & \textbf{1.67} & \textbf{2.00}\\
ALDE & $\mathbf{0.9347\pm0.0904}$ & $1.8183\pm0.1814$ & $0.8958\pm0.1099$ & 3.00 & 2.67\\
EVOLVEpro-650M & $0.7344\pm0.0966$ & $2.3676\pm0.4745$ & $0.8982\pm0.0839$ & 3.00 & 3.00\\
REAP100-650M & $0.8958\pm0.0944$ & $1.9960\pm0.2371$ & $0.8858\pm0.1177$ & 3.33 & 3.67\\
RF & $0.8294\pm0.1070$ & $1.9849\pm0.2533$ & $0.8934\pm0.1224$ & 4.00 & 3.67\\
\bottomrule
\end{tabular}
\end{table}

\subsection{Robustness across initialization cohorts}
\label{sec:cohort_sensitivity}
\begin{figure}[t]
\centering
\includegraphics[width=\linewidth]{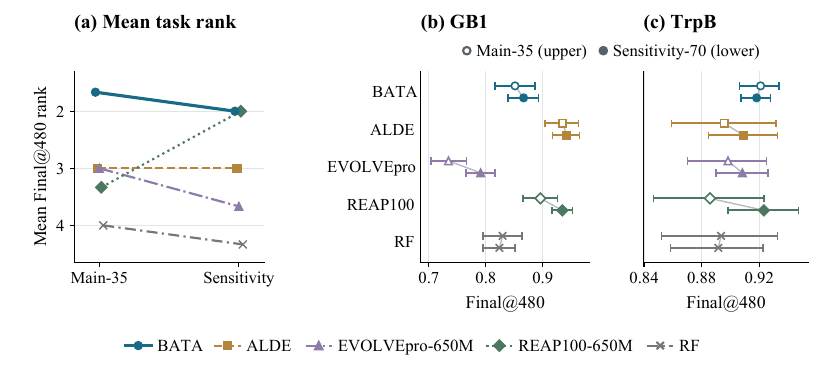}
\caption{\textbf{Initialization sensitivity.} (a) Mean Final@480 ranks in Main-35 and the sensitivity setting, which uses 70 initializations for GB1 and TrpB and retains 35 for PABP. (b,c) Mean Final@480 and 95\% bootstrap intervals for each cohort. Within each method, upper markers show Main-35 and lower markers show the separate 70-initialization cohort. Colors and marker shapes identify methods. Gray lines connect cohort means.}
\label{fig:robustness}
\end{figure}

The larger sensitivity cohort changes the ordering of methods (Figure~\ref{fig:robustness}). BATA and REAP100 share a mean Final@480 rank of 2.00 in this cohort, and ALDE still leads on GB1. On TrpB, REAP100 has a slightly higher mean Final@480 than BATA (0.9232 versus 0.9181), while BATA has higher Query-AUC (0.7911 versus 0.7687). The bootstrap intervals in Figure~\ref{fig:robustness} describe uncertainty within each cohort, not paired changes between cohorts. Appendix~\ref{app:additional_results} gives the full sensitivity table.

\subsection{How does feedback change the mixture?}
\label{sec:calibration_dynamics}
\begin{figure}[t]
\centering
\includegraphics[width=\linewidth]{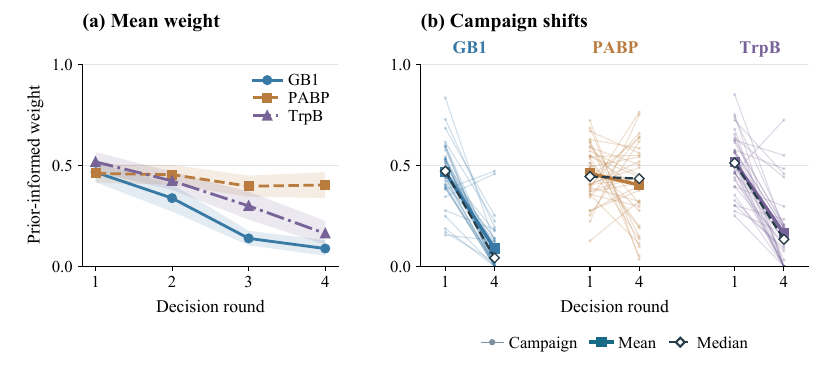}
\caption{\textbf{Feedback changes predictor weights.} (a) Mean weights on the prior-informed predictor, with pointwise 95\% bootstrap intervals over 35 Main-35 campaigns per task. (b) Light lines connect each campaign's first and last weights; squares and open diamonds show the mean and median, respectively. A lower prior-informed weight gives the task-specific predictor more influence in the combined ranking.}
\label{fig:feedback}
\end{figure}

Across the four batch decisions, the mean weight on the prior-informed predictor falls from 0.466 to 0.089 on GB1 and from 0.518 to 0.163 on TrpB (Figure~\ref{fig:feedback}). The final weight is lower than the first in 33 of 35 GB1 campaigns and 34 of 35 TrpB campaigns, so the mean decline reflects most individual runs. On PABP, the mean weight falls only from 0.460 to 0.403, with 19 campaigns showing decreases and 16 showing increases.

\subsection{Why batch-aligned tail calibration?}
\label{sec:matched_objective}
\begin{figure}[t]
\centering
\includegraphics[width=\linewidth]{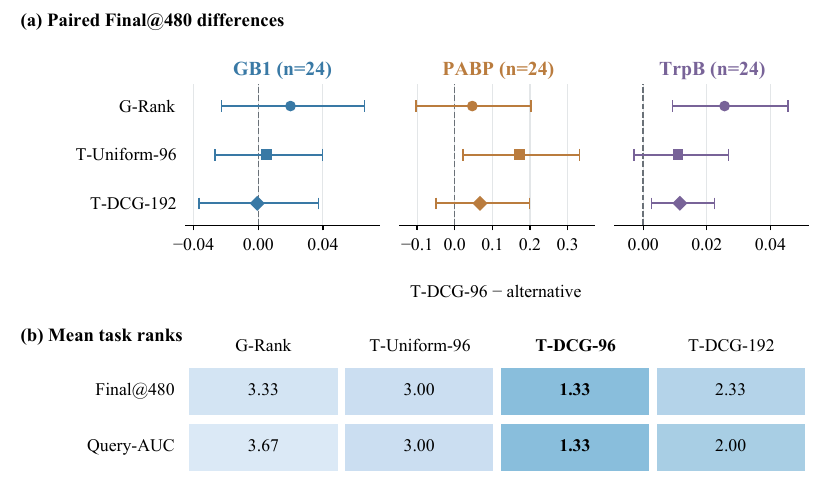}
\caption{\textbf{Calibration objectives.} (a) All nine paired Final@480 differences between T-DCG-96 (BATA) and the alternative objectives, with the study's 95\% bootstrap intervals over 24 matched groups per task. T-DCG-96 applies discounted cumulative gain (DCG) weights at a measured-fitness rank cutoff of 96. Ties use average ranks, so the number of weighted variants can differ from 96. Positive differences favor T-DCG-96; separate task axes retain the native fitness units. (b) Mean task ranks for Final@480 and Query-AUC from the same 288 optimization runs. Lower ranks are better.}
\label{fig:objective}
\end{figure}

We compare four calibration objectives: global rank error (G-Rank), uniformly weighted top-96 error (T-Uniform-96), discounted cumulative gain (DCG)-weighted top-96 error (T-DCG-96, BATA), and DCG-weighted top-192 error (T-DCG-192). The top-96 and top-192 labels denote measured-fitness rank cutoffs, using average ranks for ties. We hold the predictor configurations, out-of-fold procedure, rank combination, selection rules, initialization identities, and measurement budget fixed. The study has 24 matched groups per task and 288 optimization runs across the four objectives.
T-DCG-96 has the lowest mean task rank for both Final@480 and Query-AUC (1.33; Figure~\ref{fig:objective}b). On TrpB, its mean Final@480 exceeds G-Rank by 0.0256 (paired 95\% CI [0.0092, 0.0456]) and T-DCG-192 by 0.0115 ([0.0027, 0.0224]). On PABP, DCG weighting improves mean Final@480 over uniform top-96 weighting by 0.1719 ([0.0226, 0.3320]). GB1 has nearly equal mean Final@480 under the top-96 and top-192 objectives, with a difference of $-0.0006$. Figure~\ref{fig:objective}a shows all nine contrasts in each task's native fitness units.

\subsection{Comparison with a stronger task-specific predictor}
\label{sec:expert_sufficiency}
\begin{figure}[t]
\centering
\includegraphics[width=\linewidth]{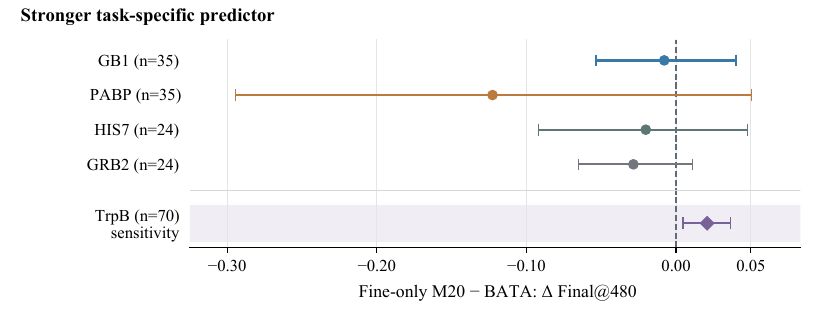}
\caption{\textbf{Stronger task-specific predictor.} Paired mean Final@480 differences (Fine-only M20 minus BATA) with 95\% confidence intervals. Positive values favor Fine-only M20. GB1 and PABP use $n=35$, HIS7 and GRB2 use $n=24$, and the separate TrpB sensitivity cohort uses $n=70$. Fine-only M20 uses a 20-member task-specific ensemble, removes the prior-informed predictor, and changes the selection rule on some tasks.}
\label{fig:expert}
\end{figure}

We compare BATA with Fine-only M20, a 20-member task-specific ensemble without the prior-informed predictor (Figure~\ref{fig:expert}). The comparison uses 35 matched pairs each for GB1 and PABP, 24 each for HIS7 and GRB2, and a separate 70-initialization sensitivity cohort for TrpB. On TrpB, Fine-only M20 reaches a mean Final@480 of 0.9388, compared with 0.9181 for BATA, giving a paired advantage of 0.0207 (95\% CI [0.0045, 0.0364]). BATA has higher means on the other four tasks, with paired intervals that include zero. Fine-only M20 changes three parts of the configuration: it increases the task-specific ensemble size, removes the prior-informed predictor, and uses Thompson-style selection on tasks where BATA uses greedy selection.
In separate controls with matched task-specific predictor capacity, BATA has a higher mean Final@480 than task-specific-only prediction on TrpB, but the paired interval includes zero (Appendix~\ref{app:expert_sufficiency}). A threshold-switching test also finds that low prior-informed weights do not consistently predict gains from switching to Fine-only M20.

\FloatBarrier

\section{Discussion and Limitations}
\label{sec:discussion}
The experiments show where feedback calibration helps. First, the prior-informed predictor's weight falls across most GB1 and TrpB campaigns, while PABP weights change less consistently. Second, the matched objective study shows task-dependent gains from fitting the high-fitness region and matching the cutoff to the batch size. Third, the stronger task-specific ensemble improves TrpB optimization in the independent 70-initialization cohort. These findings support updating the mixture as measurements accumulate, while comparing alternative predictor configurations directly. The learned weight describes agreement with the currently measured fitness ranking; it cannot by itself identify the best future predictor configuration.
Retrospective fitness landscapes let us compare methods under matched initializations and fixed measurement budgets. Prospective use must also account for laboratory constraints, assay variation, and measurement noise. BATA uses a DCA-based prior-informed predictor and estimates one mixture weight per round with five-fold OOF fitting. Therefore, its optimization performance depends on the quality of the two predictors, and its computational cost includes repeated fitting. The runtime and capacity analyses in the appendices quantify these costs and identify task-level failures.

\section{Conclusion}
BATA uses measured fitness to update two predictors and calibrate their influence on the next experimental batch. It fits a closed-form mixture weight to the observed high-fitness region, with a rank cutoff matched to the batch size. BATA achieves the best mean task rank for Final@480 in Main-35 across three protein landscapes. Matched controls show task-dependent gains from the calibration objective, while stronger-predictor comparisons identify cases that favor a task-specific ensemble. The results support updating predictor weights from experimental feedback while evaluating the predictor configuration for each task.

\label{page:main_end}
\clearpage
\section*{AI Use Statement}
Generative AI tools were used to assist literature retrieval, research ideation and experimental workflows, manuscript drafting and revision, software review, and figure preparation. All AI-assisted research outputs, code, experimental results, citations, and interpretations were reviewed and verified by the authors. All reported numerical results originate from recorded model evaluations and measured protein fitness-landscape analyses. The authors take full responsibility for the final content.
\section*{Reproducibility Statement}
Sections~\ref{sec:feedback_calibrated}--\ref{sec:bata} define the task, calibration loss, and algorithm. Section~\ref{sec:setup} and Appendix~\ref{app:reproducibility} specify the measurement budgets, predictor configurations, selection rules, and cohorts. The appendices report task-level results, paired comparisons, uncertainty estimates, and proofs. The experiment records identify the initialization sets, random seeds, query histories, and source data for the reported results.
\bibliographystyle{iclr2027_conference}
\bibliography{references}

@article{ibarraran2026efficient,
  title = {{Efficient, Few-Shot Directed Evolution with Energy Rank Alignment}},
  author = {Ibarraran, Sebastian and Chennakesavalu, Shriram and Hu, Frank and Rotskoff, Grant M.},
  journal = {Journal of Chemical Information and Modeling},
  year = {2026},
  volume = {66},
  number = {17},
  pages = {10609--10621},
  doi = {10.1021/acs.jcim.6c01333},
  url = {https://doi.org/10.1021/acs.jcim.6c01333},
}

@article{melamed2013pabp,
  title = {{Deep mutational scanning of an RRM domain of the Saccharomyces cerevisiae poly(A)-binding protein}},
  author = {Melamed, Daniel and Young, David L. and Gamble, Caitlin E. and Miller, Christina R. and Fields, Stanley},
  journal = {RNA},
  year = {2013},
  volume = {19},
  number = {11},
  pages = {1537--1551},
  doi = {10.1261/rna.040709.113},
  url = {https://doi.org/10.1261/rna.040709.113},
}

@article{morcos2011dca,
  title = {{Direct-coupling analysis of residue coevolution captures native contacts across many protein families}},
  author = {Morcos, Faruck and Pagnani, Andrea and Lunt, Bryan and Bertolino, Arianna and Marks, Debora S. and Sander, Chris and Zecchina, Riccardo and Onuchic, Jos\'e N. and Hwa, Terence and Weigt, Martin},
  journal = {Proceedings of the National Academy of Sciences},
  year = {2011},
  volume = {108},
  number = {49},
  pages = {E1293--E1301},
  doi = {10.1073/pnas.1111471108},
  url = {https://doi.org/10.1073/pnas.1111471108},
}

@article{jiang2025evolvepro,
  title = {{Rapid in silico directed evolution by a protein language model with EVOLVEpro}},
  author = {Jiang, Kaiyi and Yan, Zhaoqing and Di Bernardo, Matteo and Sgrizzi, Samantha R. and Villiger, Lukas and Kayabolen, Alisan and Kim, B. J. and Carscadden, Josephine K. and Hiraizumi, Masahiro and Nishimasu, Hiroshi and Gootenberg, Jonathan S. and Abudayyeh, Omar O.},
  journal = {Science},
  year = {2025},
  volume = {387},
  number = {6732},
  pages = {eadr6006},
  doi = {10.1126/science.adr6006},
  url = {https://doi.org/10.1126/science.adr6006},
}

@article{jarvelin2002dcg,
  title = {{Cumulated gain-based evaluation of IR techniques}},
  author = {J{\"a}rvelin, Kalervo and Kek{\"a}l{\"a}inen, Jaana},
  journal = {ACM Transactions on Information Systems},
  year = {2002},
  volume = {20},
  number = {4},
  pages = {422--446},
  doi = {10.1145/582415.582418},
  url = {https://doi.org/10.1145/582415.582418},
}

@article{wan2026evomax,
  title = {{Adaptive model-guided protein evolution with sparse data optimizes compact eukaryotic genome editors}},
  author = {Wan, Shijie and Gold, Jackson and Vure, Pranay and Mogilevsky, Casey S. and Talikoti, Ananya and Chen, Tianrong and Gupta, Aman and Biswas, Trisha and You, Zheng and Acharya, Vir and Chatterjee, Pranam and Wang, Xiao and Gao, Xue},
  journal = {Nature Biotechnology},
  year = {2026},
  doi = {10.1038/s41587-026-03272-4},
  url = {https://doi.org/10.1038/s41587-026-03272-4},
}

@article{johnston2024combinatorially,
  title = {{A combinatorially complete epistatic fitness landscape in an enzyme active site}},
  author = {Johnston, Kadina E. and Almhjell, Patrick J. and Watkins-Dulaney, Ella J. and Liu, Grace and Porter, Nicholas J. and Yang, Jason and Arnold, Frances H.},
  journal = {Proceedings of the National Academy of Sciences},
  year = {2024},
  volume = {121},
  number = {32},
  pages = {e2400439121},
  doi = {10.1073/pnas.2400439121},
  url = {https://doi.org/10.1073/pnas.2400439121},
}

@article{lin2023evolutionary,
  title = {{Evolutionary-scale prediction of atomic-level protein structure with a language model}},
  author = {Lin, Zeming and Akin, Halil and Rao, Roshan and Hie, Brian and Zhu, Zhongkai and Lu, Wenting and Smetanin, Nikita and Verkuil, Robert and Kabeli, Ori and Shmueli, Yaniv and dos Santos Costa, Allan and Fazel-Zarandi, Maryam and Sercu, Tom and Candido, Salvatore and Rives, Alexander},
  journal = {Science},
  year = {2023},
  volume = {379},
  number = {6637},
  pages = {1123--1130},
  doi = {10.1126/science.ade2574},
  url = {https://doi.org/10.1126/science.ade2574},
}

@article{russo2018tutorial,
  title = {{A Tutorial on Thompson Sampling}},
  author = {Russo, Daniel J. and Van Roy, Benjamin and Kazerouni, Abbas and Osband, Ian and Wen, Zheng},
  journal = {Foundations and Trends in Machine Learning},
  year = {2018},
  volume = {11},
  number = {1},
  pages = {1--96},
  doi = {10.1561/2200000070},
  url = {https://doi.org/10.1561/2200000070},
}

@article{ssmula2025,
  title = {{Evaluation of machine learning-assisted directed evolution across diverse combinatorial landscapes}},
  author = {Li, Francesca-Zhoufan and Yang, Jason and Johnston, Kadina E. and G{\"u}rsoy, Emre and Yue, Yisong and Arnold, Frances H.},
  journal = {Cell Systems},
  year = {2025},
  volume = {16},
  number = {9},
  pages = {101387},
  doi = {10.1016/j.cels.2025.101387},
  url = {https://doi.org/10.1016/j.cels.2025.101387},
}

@article{wu2016adaptation,
  title = {{Adaptation in protein fitness landscapes is facilitated by indirect paths}},
  author = {Wu, Nicholas C and Dai, Lei and Olson, C Anders and Lloyd-Smith, James O and Sun, Ren},
  journal = {eLife},
  year = {2016},
  volume = {5},
  pages = {e16965},
  doi = {10.7554/elife.16965},
  url = {https://doi.org/10.7554/elife.16965},
}

@article{wu2019machine,
  title = {{Machine learning-assisted directed protein evolution with combinatorial libraries}},
  author = {Wu, Zachary and Kan, S. B. Jennifer and Lewis, Russell D. and Wittmann, Bruce J. and Arnold, Frances H.},
  journal = {Proceedings of the National Academy of Sciences},
  year = {2019},
  volume = {116},
  number = {18},
  pages = {8852--8858},
  doi = {10.1073/pnas.1901979116},
  url = {https://doi.org/10.1073/pnas.1901979116},
}

@article{yang2019machine,
  title = {{Machine-learning-guided directed evolution for protein engineering}},
  author = {Yang, Kevin K. and Wu, Zachary and Arnold, Frances H.},
  journal = {Nature Methods},
  year = {2019},
  volume = {16},
  number = {8},
  pages = {687--694},
  doi = {10.1038/s41592-019-0496-6},
  url = {https://doi.org/10.1038/s41592-019-0496-6},
}

@article{yang2025active,
  title = {{Active learning-assisted directed evolution}},
  author = {Yang, Jason and Lal, Ravi G. and Bowden, James C. and Astudillo, Raul and Hameedi, Mikhail A. and Kaur, Sukhvinder and Hill, Matthew and Yue, Yisong and Arnold, Frances H.},
  journal = {Nature Communications},
  year = {2025},
  volume = {16},
  number = {1},
  pages = {714},
  doi = {10.1038/s41467-025-55987-8},
  url = {https://doi.org/10.1038/s41467-025-55987-8},
}

@inproceedings{prospero2025,
  title = {{ProSpero: Active Learning for Robust Protein Design Beyond Wild-Type Neighborhoods}},
  author = {Kmicikiewicz, Michal and Fortuin, Vincent and Szczurek, Ewa},
  booktitle = {NeurIPS},
  year = {2025},
  url = {https://proceedings.neurips.cc/paper_files/paper/2025/hash/bde1c75d7d89e64d6d93dce1c68dc3d8-Abstract-Conference.html},
  volume = {38},
  pages = {131015--131048},
  doi = {10.52202/085713-4364},
}

@inproceedings{notin2023proteingym,
  title = {{ProteinGym: Large-Scale Benchmarks for Protein Fitness Prediction and Design}},
  author = {Notin, Pascal and Kollasch, Aaron and Ritter, Daniel and van Niekerk, Lood and Paul, Steffanie and Spinner, Han and Rollins, Nathan and Shaw, Ada and Orenbuch, Rose and Weitzman, Ruben and Frazer, Jonathan and Dias, Mafalda and Franceschi, Dinko and Gal, Yarin and Marks, Debora},
  booktitle = {NeurIPS},
  year = {2023},
  url = {https://proceedings.neurips.cc/paper_files/paper/2023/hash/cac723e5ff29f65e3fcbb0739ae91bee-Abstract-Datasets_and_Benchmarks.html},
  volume = {36},
  pages = {64331--64379},
  doi = {10.52202/075280-2810},
}

@article{xu2026rank,
  title = {{Rank-guided learning accelerates automated enzyme engineering}},
  author = {Xu, Jingyi and Zheng, Yan and Sundarraj, Rajamanikandan and Woycechowsky, Kenneth and Yuchi, Zhiguang and Yuan, Yingjin},
  journal = {Nature Communications},
  year = {2026},
  volume = {17},
  pages = {9584},
  doi = {10.1038/s41467-026-76264-2},
  url = {https://doi.org/10.1038/s41467-026-76264-2},
}

@inproceedings{bal2025gameopt,
  title = {Optimistic Games for Combinatorial {Bayesian} Optimization with Application to Protein Design},
  author = {Bal, Melis Ilayda and Sessa, Pier Giuseppe and Mutny, Mojmir and Krause, Andreas},
  booktitle = {ICLR},
  year = {2025},
  url = {https://proceedings.iclr.cc/paper_files/paper/2025/hash/434743b59a2ef42d657e61e06f0dd97b-Abstract-Conference.html}
}

@inproceedings{amin2025clonebo,
  title = {{Bayesian} Optimization of Antibodies Informed by a Generative Model of Evolving Sequences},
  author = {Amin, Alan and Gruver, Nate and Kuang, Yilun and Li, Yucen and Elliott, Hunter and McCarter, Calvin and Raghu, Aniruddh and Greenside, Peyton and Wilson, Andrew Gordon},
  booktitle = {ICLR},
  year = {2025},
  url = {https://proceedings.iclr.cc/paper_files/paper/2025/hash/6cdd4ce9330025967dd1ed0bed3010f5-Abstract-Conference.html}
}

@inproceedings{karimi2025distillation,
  title = {Data Distillation for Extrapolative Protein Design through Exact Preference Optimization},
  author = {Karimi, Mostafa and Banerjee, Sharmi and Jaakkola, Tommi and Dubrov, Bella and Shang, Shang and Benson, Ron},
  booktitle = {ICLR},
  year = {2025},
  url = {https://proceedings.iclr.cc/paper_files/paper/2025/hash/a9ea92ef18aae17627d133534209e640-Abstract-Conference.html}
}

@inproceedings{kirjner2024ggs,
  title = {Improving Protein Optimization with Smoothed Fitness Landscapes},
  author = {Kirjner, Andrew and Yim, Jason and Samusevich, Raman and Bracha, Shahar and Jaakkola, Tommi and Barzilay, Regina and Fiete, Ila},
  booktitle = {ICLR},
  year = {2024},
  url = {https://proceedings.iclr.cc/paper_files/paper/2024/hash/cbb7a23649b25001e797a726cf75498e-Abstract-Conference.html}
}

@inproceedings{ghaffari2024robust,
  title = {Robust Model-Based Optimization for Challenging Fitness Landscapes},
  author = {Ghaffari, Saba and Saleh, Ehsan and Schwing, Alex and Wang, Yu-Xiong and Burke, Martin and Sinha, Saurabh},
  booktitle = {ICLR},
  year = {2024},
  url = {https://proceedings.iclr.cc/paper_files/paper/2024/hash/f84ceafb242f0de36f8c49452fbae6de-Abstract-Conference.html}
}

@inproceedings{bogensperger2025vlgpo,
  title = {A Variational Perspective on Generative Protein Fitness Optimization},
  author = {Bogensperger, Lea and Narnhofer, Dominik and Allam, Ahmed and Schindler, Konrad and Krauthammer, Michael},
  booktitle = {ICML},
  series = {Proceedings of Machine Learning Research},
  volume = {267},
  pages = {4700--4712},
  publisher = {PMLR},
  year = {2025},
  url = {https://proceedings.mlr.press/v267/bogensperger25a.html}
}

@inproceedings{lee2024latprotrl,
  title = {Robust Optimization in Protein Fitness Landscapes Using Reinforcement Learning in Latent Space},
  author = {Lee, Minji and Vecchietti, Luiz Felipe and Jung, Hyunkyu and Ro, Hyun Joo and Cha, Meeyoung and Kim, Ho Min},
  booktitle = {ICML},
  series = {Proceedings of Machine Learning Research},
  volume = {235},
  pages = {26976--26990},
  publisher = {PMLR},
  year = {2024},
  url = {https://proceedings.mlr.press/v235/lee24x.html}
}

@inproceedings{wang2024knowrlm,
  title = {Knowledge-aware Reinforced Language Models for Protein Directed Evolution},
  author = {Wang, Yuhao and Zhang, Qiang and Qin, Ming and Zhuang, Xiang and Li, Xiaotong and Gong, Zhichen and Wang, Zeyuan and Zhao, Yu and Yao, Jianhua and Ding, Keyan and Chen, Huajun},
  booktitle = {ICML},
  series = {Proceedings of Machine Learning Research},
  volume = {235},
  pages = {52260--52273},
  publisher = {PMLR},
  year = {2024},
  url = {https://proceedings.mlr.press/v235/wang24cq.html}
}

@inproceedings{qu2026dynmeanbo,
  title = {Incorporating Expert Priors into {Bayesian} Optimization via Dynamic Mean Decay},
  author = {Qu, Chongqi and Liu, Meiqin and Lan, Jian and Dong, Shanling and Liu, Zhunga},
  booktitle = {ICLR},
  year = {2026},
  url = {https://proceedings.iclr.cc/paper_files/paper/2026/hash/020e313d40a7c060ed07a10cef287750-Abstract-Conference.html}
}

@inproceedings{menet2026tosfit,
  title = {{Thompson} Sampling via Fine-Tuning of {LLMs}},
  author = {Menet, Nicolas and Terzic, Aleksandar and Hersche, Michael and Krause, Andreas and Rahimi, Abbas},
  booktitle = {ICLR},
  year = {2026},
  url = {https://proceedings.iclr.cc/paper_files/paper/2026/hash/707a2d58641b2192203b4bf4c532cfe1-Abstract-Conference.html}
}

@inproceedings{zeng2026bolt,
  title = {Scaling Multi-Task {Bayesian} Optimization with Large Language Models},
  author = {Zeng, Yimeng and Maus, Natalie and Jones, Haydn and Tao, Jeffrey and Wan, Fangping and Der Torossian Torres, Marcelo and de la Fuente, Cesar and Marcus, Ryan and Bastani, Osbert and Gardner, Jacob},
  booktitle = {ICLR},
  year = {2026},
  url = {https://proceedings.iclr.cc/paper_files/paper/2026/hash/a4a3689e5ca538e340f58cdd29f513a9-Abstract-Conference.html}
}

@article{freschlin2022navigate,
  title = {Machine learning to navigate fitness landscapes for protein engineering},
  author = {Freschlin, Chase R. and Fahlberg, Sarah A. and Romero, Philip A.},
  journal = {Current Opinion in Biotechnology},
  volume = {75},
  pages = {102713},
  year = {2022},
  doi = {10.1016/j.copbio.2022.102713}
}

@article{notin2024functional,
  title = {Machine learning for functional protein design},
  author = {Notin, Pascal and Rollins, Nathan and Gal, Yarin and Sander, Chris and Marks, Debora},
  journal = {Nature Biotechnology},
  volume = {42},
  number = {2},
  pages = {216--228},
  year = {2024},
  doi = {10.1038/s41587-024-02127-0}
}

@article{listov2024opportunities,
  title = {Opportunities and challenges in design and optimization of protein function},
  author = {Listov, Dina and Goverde, Casper A. and Correia, Bruno E. and Fleishman, Sarel Jacob},
  journal = {Nature Reviews Molecular Cell Biology},
  volume = {25},
  number = {8},
  pages = {639--653},
  year = {2024},
  doi = {10.1038/s41580-024-00718-y}
}

@article{yang2024opportunities,
  title = {Opportunities and Challenges for Machine Learning-Assisted Enzyme Engineering},
  author = {Yang, Jason and Li, Francesca-Zhoufan and Arnold, Frances H.},
  journal = {ACS Central Science},
  volume = {10},
  number = {2},
  pages = {226--241},
  year = {2024},
  doi = {10.1021/acscentsci.3c01275}
}

@article{rapp2024selfdriving,
  title = {Self-driving laboratories to autonomously navigate the protein fitness landscape},
  author = {Rapp, Jacob T. and Bremer, Bennett J. and Romero, Philip A.},
  journal = {Nature Chemical Engineering},
  volume = {1},
  number = {1},
  pages = {97--107},
  year = {2024},
  doi = {10.1038/s44286-023-00002-4}
}

@article{sandhu2025fitness,
  title = {Computational and Experimental Exploration of Protein Fitness Landscapes: Navigating Smooth and Rugged Terrains},
  author = {Sandhu, Mahakaran and Chen, John Z. and Matthews, Dana S. and Spence, Matthew A. and Pulsford, Sacha B. and Gall, Barnabas and Kaczmarski, Joe A. and Nichols, James and Tokuriki, Nobuhiko and Jackson, Colin J.},
  journal = {Biochemistry},
  volume = {64},
  number = {8},
  pages = {1673--1684},
  year = {2025},
  doi = {10.1021/acs.biochem.4c00673}
}

@article{herrera2025bestpractices,
  title = {Best Practices for Machine Learning-Assisted Protein Engineering},
  author = {Herrera-Rocha, Fabio and Medina-Ortiz, David and Mauz, Fabian and Pleiss, Juergen and Davari, Mehdi D.},
  journal = {Journal of Chemical Information and Modeling},
  volume = {65},
  number = {23},
  pages = {12655--12667},
  year = {2025},
  doi = {10.1021/acs.jcim.5c01983}
}
\clearpage
\appendix
\setcounter{figure}{0}
\renewcommand{\thefigure}{A\arabic{figure}}
\renewcommand{\theHfigure}{appendix.\arabic{figure}}
\setcounter{table}{0}
\renewcommand{\thetable}{A\arabic{table}}
\renewcommand{\theHtable}{appendix.\arabic{table}}
\section{Reproducibility and Protocol Details}
\label{app:reproducibility}

\subsection{Measurements and metrics}
Each optimization campaign starts with 96 measured variants and selects four batches of 96 distinct variants. The simulator reveals a batch's fitness values only after every candidate in that batch has been selected. Previously measured variants are excluded from later batches. Model fitting, cross-validation, weight estimation, and selection use only the currently revealed labels. A complete campaign therefore contains 480 unique measurements.
Let $b_q$ be the best fitness observed after $q$ queries. Final@480 is $b_{480}$. Query-AUC measures the area under the best-so-far fitness curve:
\begin{equation}
\mathrm{Query\mbox{-}AUC}=\frac{b_{96}+2(b_{192}+b_{288}+b_{384})+b_{480}}{8}.
\end{equation}
This is the normalized trapezoidal area from queries 96--480. Each campaign gives one Final value and one Query-AUC value. The plotted lines connect batch endpoints; no within-batch ordering of outcomes is reconstructed.

\subsection{Sequence data and predictors}
Each task keeps the candidate ordering and fitness scale from its data release. PABP, HIS7, and GRB2 use the frozen ProteinGym v1.3 substitution tables \citep{notin2023proteingym}. GB1 uses the scale-to-maximum table from the energy-rank-alignment data release, and TrpB uses the fitness table distributed with ALDE. Before replay, input identities are checked against the experiment manifests. The reported comparisons apply no additional filtering based on fitness outcomes.
The prior-informed predictor contains five Ridge models fitted to direct coupling analysis (DCA) features, which summarize evolutionary sequence statistics. Ridge fitting standardizes features and fitness using only its training subset. The regularization parameter is selected from 17 logarithmically spaced values between $10^{-4}$ and $10^4$, followed by fitting five bootstrap models. The task-specific predictors use Ridge regression or random forest (RF) models, as listed in Table~\ref{tab:predictors}. ESM2-650M provides fixed sequence representations \citep{lin2023evolutionary}; the predictor fitted to these representations is updated after each revealed batch.

\begin{table}[!htbp]
\centering\small
\caption{Task-specific predictors and selection rules in the main BATA comparison. Every task uses five DCA-based Ridge models for the prior-informed predictor.}
\label{tab:predictors}
\begin{tabular}{llll}\toprule
Task & Representation & Task-specific predictor & Selection\\\midrule
GB1 & One-hot & Five-member Ridge ensemble & Thompson-style\\
PABP & ESM2-650M & One random forest, 100 trees & Greedy mean\\
TrpB & ESM2-650M & Five bootstrap forests, 100 trees each & Thompson-style\\\bottomrule
\end{tabular}
\end{table}
Five-fold out-of-fold (OOF) fitting starts from a seeded permutation of the measured indices. Each fold model is trained on the other four folds, so a variant's OOF prediction comes from models fitted without its fitness value. Ensemble predictions are averaged before forming each predictor's OOF rank vector. For selection from the unmeasured pool, each fitted ensemble member is ranked separately. PABP reuses its single task-specific prediction vector across the five prior-informed ensemble members. Greedy selection orders candidates by their mean combined score. Thompson-style selection samples a member for each choice and skips candidates already selected. Candidate-index order breaks selection ties deterministically.
Percentile ranks are average ascending ranks divided by the vector length. The descending fitness ranks in Eq.~\ref{eq:tail_weights} also use average ranks. Therefore, a tied group at the cutoff can change how many observations receive nonzero weight. The cutoff remains $\min(B,n_t)$. In Eq.~\ref{eq:bata_weight}, the implementation sets the prior-informed predictor's weight to zero when the denominator is no greater than float64 machine epsilon. Proposition~\ref{prop:bata_optimality} states the exact mathematical result for a zero denominator.

\subsection{Separate evaluation cohorts}
Main-35 contains 525 campaigns: five methods on three tasks, each with 35 matched initializations. The larger sensitivity setting adds 700 campaign records for the five methods on GB1 and TrpB, using 70 initializations per task; PABP keeps its original 35. Campaigns are matched within each task and cohort. Therefore, the lines between cohorts in Figure~\ref{fig:robustness} compare cohort summaries. They are not paired campaign differences. Method means are ranked within each task, and these ranks are then averaged across tasks.
The calibration-objective comparison contains 24 matched groups per task and four objectives, giving 288 optimization runs. Predictor configurations, five-fold OOF fitting, rank combination, selection rules, initialization identities, and measurement budgets are fixed. The 72 reused T-DCG-96 trajectories use the rank cutoff of 96 and DCG weights defined in Eq.~\ref{eq:tail_weights}. Reuse followed exact checks that the candidate trajectories matched the current configuration. The controls with matched task-specific predictor capacity, the stronger 20-member task-specific ensemble (Fine-only M20), and the threshold-switching rule G20 each use separate cohorts. G20 switches to Fine-only M20 when the prior-informed predictor's weight falls below 0.20.

\subsection{Uncertainty and reproducibility checks}
Curve bands are pointwise 95\% confidence intervals (CIs) from percentile-bootstrap resampling of complete campaigns. Paired contrasts resample matched campaign differences within each task. Figure~\ref{fig:objective} uses the released intervals for the calibration-objective study. The controls with matched task-specific predictor capacity use the prespecified 20,000-resample calculation. Table standard deviations describe variation among campaigns; CIs describe uncertainty in a mean or mean difference. Paired comparisons use matched differences, not overlap between pointwise curve bands.
The public code repository at \url{https://github.com/John-Lin98/BATA} provides initialization identities, algorithm seeds, result tables, and scripts for the reported analyses. The figure data identify their task and cohort, allowing means, paired differences, and batch trajectories to be checked independently. Assay data, alignments, pretrained weights, and feature caches are not bundled. The code documentation specifies external sources, fixed revisions, input hashes, and feature-rebuilding commands. The redistribution terms for the GB1 and TrpB alignments remain unconfirmed; these alignments are obtained separately from the identified provider.
\FloatBarrier
\section{Additional Experimental Results}
\label{app:additional_results}

\subsection{Full main and sensitivity results}
\begin{figure}[!htbp]
\centering\includegraphics[width=\linewidth]{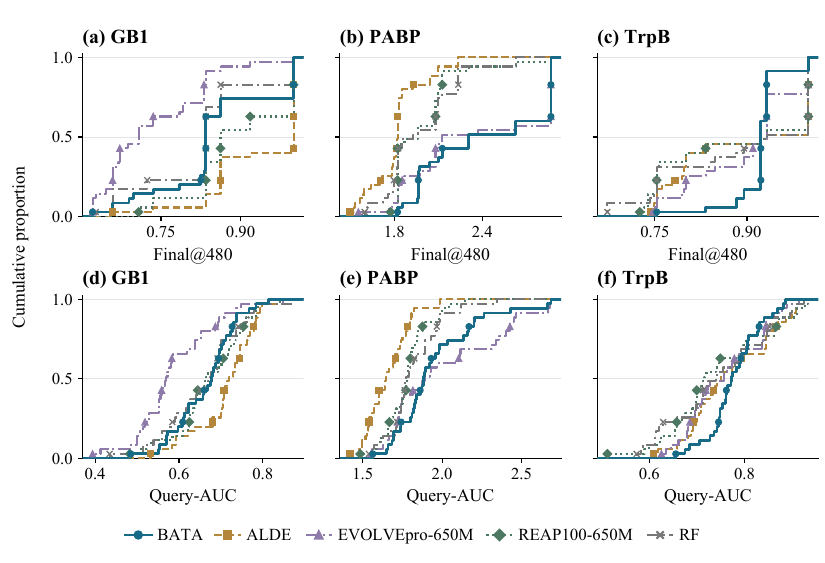}
\caption{\textbf{Campaign outcome distributions.} Empirical cumulative distributions of Main-35 Final@480 and Query-AUC. Each curve contains 35 outcomes for one method and task, measured on that task's native fitness scale.}
\label{fig:distributions}
\end{figure}

Table~\ref{tab:main_performance} reports Main-35 endpoints, and Table~\ref{tab:main_auc} reports Query-AUC for the same campaigns. Figure~\ref{fig:distributions} shows the full campaign distributions. Each empirical cumulative curve includes all 35 outcomes for one method and task, showing the variation behind the reported mean.

\begin{table}[!htbp]
\centering\small\setlength{\tabcolsep}{3pt}
\caption{\textbf{Main-35 discovery speed.} Query-AUC means $\pm$ sample standard deviation (SD) over 35 matched initializations per task. Higher Query-AUC and lower mean rank are better.}
\label{tab:main_auc}
\begin{tabular}{lcccc}\toprule
Method & GB1 & PABP & TrpB & Mean rank\\\midrule
BATA & $0.6638\pm0.0727$ & $1.9530\pm0.2650$ & $\mathbf{0.7783\pm0.0553}$ & \textbf{2.00}\\
ALDE & $\mathbf{0.7111\pm0.0736}$ & $1.6602\pm0.1351$ & $0.7689\pm0.0888$ & 2.67\\
EVOLVEpro-650M & $0.5877\pm0.0945$ & $\mathbf{2.0122\pm0.3526}$ & $0.7579\pm0.0808$ & 3.00\\
REAP100-650M & $0.6723\pm0.0847$ & $1.7809\pm0.1391$ & $0.7426\pm0.1056$ & 3.67\\
RF & $0.6600\pm0.0920$ & $1.8190\pm0.1699$ & $0.7449\pm0.1035$ & 3.67\\\bottomrule
\end{tabular}
\end{table}

Table~\ref{tab:sensitivity70} reports both metrics in the separate sensitivity setting. GB1 and TrpB use 70 initializations each, while PABP retains Main-35. BATA and REAP100 tie for the lowest Final mean rank, and BATA has the lowest Query-AUC mean rank. These rankings compare cohort summaries; campaign pairs are matched only within a cohort.
\begin{table}[!htbp]
\centering
\caption{\textbf{Sensitivity results.} Means $\pm$ standard deviation (SD) and mean task ranks for GB1 and TrpB with 70 initializations and PABP with 35. The two sections report Final@480 and Query-AUC, respectively.}
\label{tab:sensitivity70}
\includegraphics[width=\linewidth]{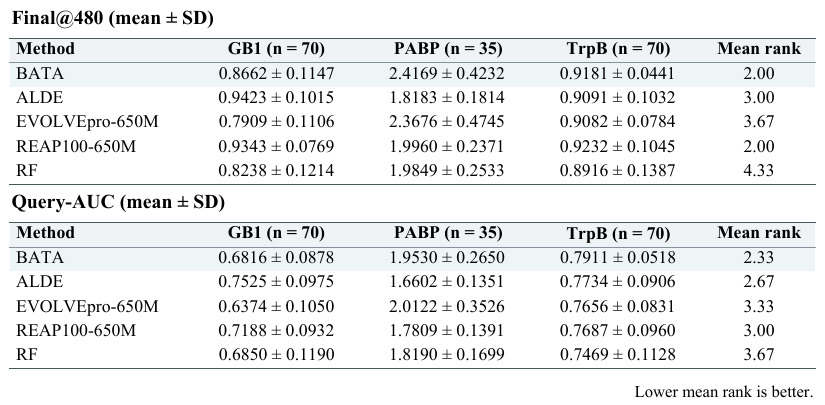}
\end{table}

\FloatBarrier
\subsection{Calibration outcomes and weight variation}
\begin{figure}[!htbp]
\centering\includegraphics[width=\linewidth]{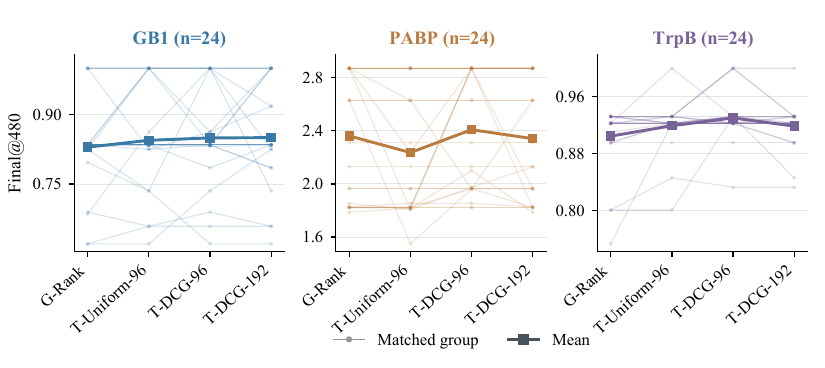}
\caption{\textbf{Matched groups across calibration objectives.} Each light line connects one group's outcomes under the four objectives; squares show objective means. Each task contains 24 matched groups.}
\label{fig:objective_groups}
\end{figure}

The four-objective comparison changes the calibration loss while keeping the predictors and selection rules fixed. Table~\ref{tab:matched_objective} reports task means, and Figure~\ref{fig:objective_groups} shows all 24 matched groups. Figure~\ref{fig:objective}a reports all nine paired differences with their intervals, including those that span zero.

\begin{table}[!htbp]\centering\small
\caption{\textbf{Calibration-objective results.} Mean Final@480 over 24 matched groups per task. Final and Query-AUC ranks are computed from task means and then averaged; lower ranks are better.}
\label{tab:matched_objective}
\begin{tabular}{lccccc}\toprule
Objective & GB1 & PABP & TrpB & Final rank & AUC rank\\\midrule
G-Rank & 0.8297 & 2.3605 & 0.9045 & 3.33 & 3.67\\
T-Uniform-96 & 0.8445 & 2.2353 & 0.9191 & 3.00 & 3.00\\
T-DCG-96 & 0.8496 & \textbf{2.4072} & \textbf{0.9301} & \textbf{1.33} & \textbf{1.33}\\
T-DCG-192 & \textbf{0.8503} & 2.3401 & 0.9186 & 2.33 & 2.00\\\bottomrule
\end{tabular}
\end{table}

\begin{figure}[!htbp]
\centering\includegraphics[width=\linewidth]{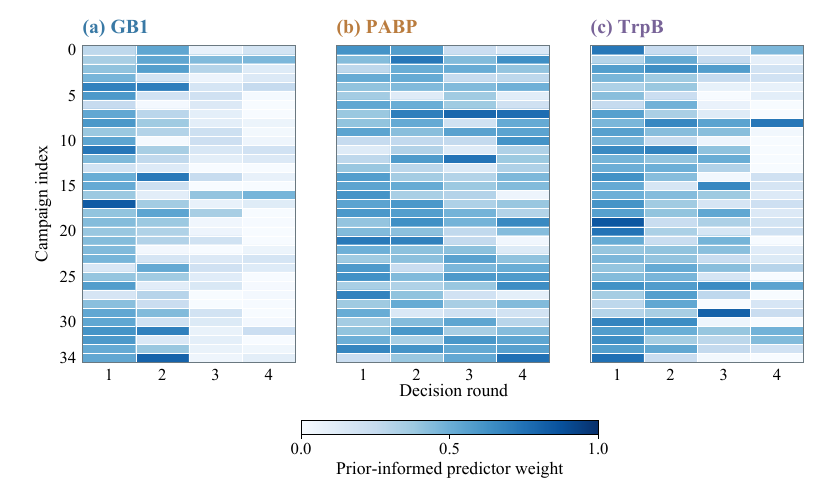}
\caption{\textbf{Mixture weights across all campaigns.} Each task contains 35 campaigns and four decision rounds. Zero assigns all weight to the task-specific predictor; one assigns all weight to the prior-informed predictor. All heatmaps use this shared scale.}
\label{fig:weight_heatmaps}
\end{figure}

Figure~\ref{fig:weight_heatmaps} shows all 420 mixture weights in the original campaign order. GB1 and TrpB mainly shift toward the task-specific predictor, while PABP shows more varied changes. Campaign order is kept fixed across rounds and does not depend on final fitness.

\FloatBarrier
\subsection{Runtime, capacity, and task-level limits}
\begin{figure}[!htbp]
\centering\includegraphics[width=\linewidth]{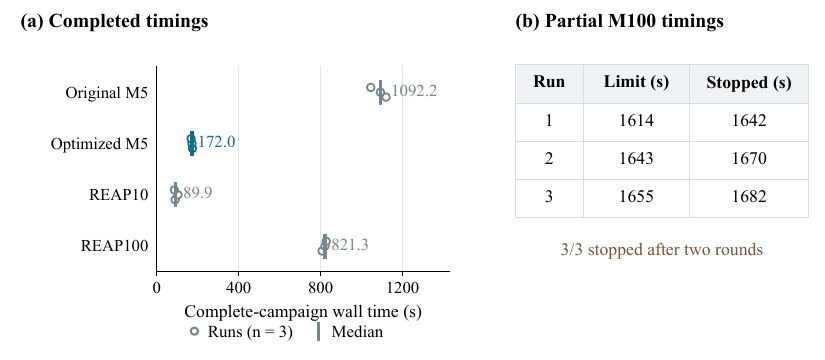}
\caption{\textbf{Runtime and ensemble size.} (a) Complete-campaign wall times and medians from three engineering timings per configuration. (b) Elapsed times of three M100 attempts stopped after two rounds at their time limits. These times cover incomplete campaigns.}
\label{fig:runtime}
\end{figure}

An implementation optimization reduced median warm-cache runtime for the five-member (M5) configuration from 1092.19 to 171.96 seconds, with identical recorded candidate trajectories and metrics. Optimized M5 took 1.91 times the REAP10 runtime under this timing protocol. Figure~\ref{fig:runtime}a shows the three completed timings for each configuration. The 100-member (M100) runs in panel (b) reached their declared time limits after two rounds, so their elapsed times cover only part of a campaign.
With 24 matched initializations, BATA using ESM2-650M reached mean Final values of 2.4072 on PABP and 0.9301 on TrpB. The evaluated EVOLVEpro-15B adaptation reached 2.1177 and 0.8257, respectively. Increasing the BATA ensemble to 20 members (M20) increased runtime and still gave a lower Final than REAP100 on the complete TrpB development cohort. Each comparison evaluates the full predictor and selection configuration, so language-model size is one of several differences.
Additional studies show where other configurations perform better. The global mean squared error (Global-MSE) configuration outperformed BATA on PABP in the 24-initialization auxiliary study. FolDE led under the separate 48-query PABP protocol, and the HIS7 and GRB2 transfer comparisons favored RF and REAP100, respectively. Global-MSE uses raw-score fusion and also changes covariance estimation and cross-fitting; the four-objective study changes only the rank-space calibration loss. These studies use separate protocols and cohorts and are reported independently of Main-35.
\FloatBarrier
\section{Expert Comparisons and Switching Diagnostics}
\label{app:expert_sufficiency}

\subsection{Stronger task-specific prediction}
Fine-only M20 uses 20 task-specific ensemble members and removes the prior-informed predictor. Figure~\ref{fig:expert} reports 188 matched pairs: 35 each for GB1 and PABP, 24 each for HIS7 and GRB2, and 70 from a separate TrpB sensitivity cohort. On TrpB, mean Final is 0.9388 for Fine-only M20 and 0.9181 for BATA. The paired advantage of Fine-only M20 is 0.0207, with a 95\% confidence interval (CI) of [0.0045, 0.0364]. These TrpB initializations are independent of Main-35.
Fine-only M20 changes ensemble capacity and uses Thompson-style selection on every task, including PABP and GRB2 where BATA uses greedy selection. Therefore, the comparison evaluates these configuration changes along with removal of the prior-informed predictor. The following controls keep the original task-specific predictor capacity and selection rule.

\subsection{Controls with matched task-specific predictor capacity}
\begin{figure}[!ht]
\centering
\includegraphics[width=\linewidth]{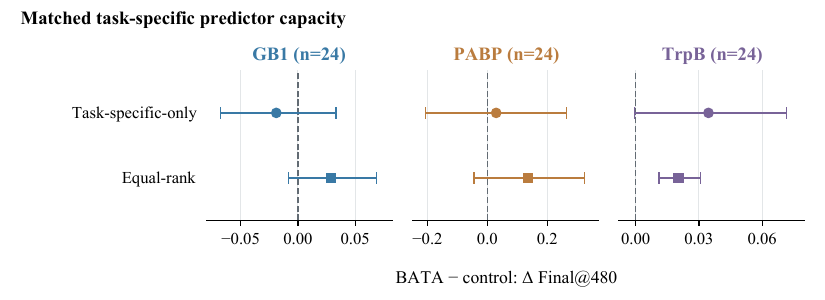}
\caption{\textbf{Controls with matched task-specific predictor capacity.} Paired mean Final@480 differences and 95\% bootstrap intervals for BATA minus task-specific-only prediction and equal-rank combination, using 24 matched initializations per main task. Each task has a separate axis in its native fitness units. Positive values favor BATA.}
\label{fig:capacity_controls}
\end{figure}

The control set contains 216 runs: BATA, task-specific-only prediction, and equal-rank combination, each evaluated on 24 matched initializations per main task. The task-specific predictors keep their original configurations: five Ridge models on GB1, five bootstrap random forests on TrpB, and one random forest with 100 trees on PABP. PABP task-specific-only predictions reuse the released EVOLVEpro/RF100 control. Figure~\ref{fig:capacity_controls} reports BATA minus control, so positive values favor BATA. Figure~\ref{fig:expert} uses Fine-only M20 minus BATA.

On TrpB, BATA exceeds task-specific-only prediction by a mean of 0.0345306. The prespecified 20,000-resample CI is [$-0.000608$, 0.071601], so uncertainty in this mean advantage extends across zero. Compared with equal-rank combination, BATA has a mean advantage of 0.020349 and a CI of [0.011045, 0.030704]. The independent TrpB-70 M20 comparison evaluates a larger task-specific ensemble, so it answers a different question from these controls.

\subsection{Threshold development and independent confirmation}
The 0.20-threshold switching rule (G20) uses Fine-only M20 when BATA's fitted prior-informed predictor weight is below 0.20. Otherwise, it keeps the BATA combination. Development evaluated thresholds 0.20, 0.30, and 0.40 and Fine-only M20 with five seeds per task across five tasks. These four variants required 100 new runs; the comparison also reused 25 BATA controls. G20 had the lowest development macro normalized regret: 0.1040, compared with 0.5040 and 0.5495 for the higher thresholds. The threshold was then fixed before generating a separate confirmation cohort.
During development, G20 switched in 19 of 100 decisions, while G30 (threshold 0.30) and G40 (0.40) switched in 32 and 45, respectively. These more frequent switching rules had worse development scores. Confirmation used ten new initializations per task for BATA, Fine-only M20, and G20, giving 150 complete 480-query campaigns (Table~\ref{tab:g20_fresh}). The confirmation initialization identities and algorithm seeds have no recorded overlap with the historical manifest bank.

\begin{figure}[!ht]
\centering\includegraphics[width=\linewidth]{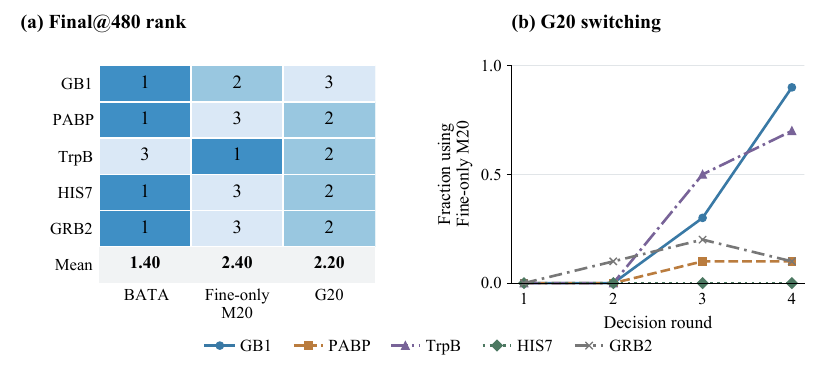}
\caption{\textbf{Switching-rule results.} (a) Task-wise Final ranks and released mean ranks in the independent ten-initialization cohort. (b) Fraction of campaigns using Fine-only M20 at each round. G20 uses the fixed threshold of 0.20. Decision rounds within a campaign are dependent.}
\label{fig:g20}
\end{figure}

\begin{table}[!ht]\centering\small
\caption{\textbf{Independent confirmation of the switching rule.} Mean Final@480 over ten matched initializations per task. Bold marks the best task mean. Ranks use the released deterministic tie rule: decreasing mean, then increasing arm identifier.}
\label{tab:g20_fresh}
\begin{tabular}{lccc}\toprule
Task & BATA & G20 & Fine-only M20\\\midrule
GB1 & \textbf{0.8802} & 0.8519 & 0.8549\\
PABP & \textbf{2.3132} & 2.2571 & 2.0830\\
TrpB & 0.9399 & 0.9554 & \textbf{0.9665}\\
HIS7 & \textbf{1.4138} & \textbf{1.4138} & 1.3227\\
GRB2 & \textbf{0.6276} & \textbf{0.6276} & 0.5964\\\midrule
Mean rank & \textbf{1.40} & 2.20 & 2.40\\\bottomrule
\end{tabular}
\end{table}
In confirmation, G20 used Fine-only M20 in 30\% of decision rounds on GB1 and TrpB, 5\% on PABP, 10\% on GRB2, and none on HIS7. Its paired Final differences from BATA were $-0.0284$ on GB1 (95\% CI [$-0.0731$, 0]), $-0.0560$ on PABP ([$-0.1681$, 0]), $+0.0155$ on TrpB ([0, 0.0359]), and zero on HIS7 and GRB2. Figure~\ref{fig:g20} shows the ranks and switching frequency at each round. A low prior-informed predictor weight describes the current mixture but does not consistently predict gains from switching to the larger task-specific ensemble.

\FloatBarrier

\section{Proofs for the Calibration Rule}
\label{app:calibration_properties}
Both propositions analyze one round with fixed fitted predictions, ranks, and sample weights. Their conclusions apply to the calibration objective in Eq.~\ref{eq:bata_objective}.

\subsection{Proof of Proposition~\ref{prop:bata_optimality}}
Write $A=\sum_iq_id_i^2$, $C=\sum_iq_id_iz_i$, and $K=\sum_iq_iz_i^2$. The loss expands to
\begin{equation}
\mathcal L_{\rm BATA}(w)=Aw^2-2Cw+K,\qquad A\ge0.
\end{equation}
For $A>0$, completing the square gives
\begin{equation}
\mathcal L_{\rm BATA}(w)=A\left(w-\frac CA\right)^2+K-\frac{C^2}{A}.
\end{equation}
The unique unconstrained minimizer is $C/A$. Over $[0,1]$, the minimizer is the closest point to $C/A$ in that interval, $\Pi_{[0,1]}(C/A)$. If $A=0$, every term with $q_i>0$ has $d_i=0$. Hence $C=0$, and the loss is constant. The convention $w=0$ is then optimal. The minimum loss is at most the loss at either endpoint, each of which uses one predictor alone.
The implementation sets $w=0$ when $A$ is no larger than float64 machine epsilon, treating an almost-flat objective as numerically degenerate. The exact uniqueness claim in the proposition applies to the mathematical case $A>0$.

\subsection{Proof of Proposition~\ref{prop:bata_invariance}}
A strictly increasing transformation preserves order and ties: $u<v$ if and only if $h(u)<h(v)$, and $u=v$ if and only if $h(u)=h(v)$. Therefore, average ranks and their percentile normalization stay unchanged. Transforming either fixed, ensemble-averaged OOF score vector leaves $r^p,r^a,d,z$ and the weight in Eq.~\ref{eq:bata_weight} unchanged. The observed fitness ranks and sample weights $q_i$ also stay unchanged.
Applying the same argument to each member's fixed scores in the unmeasured pool leaves both percentile-rank vectors in Eq.~\ref{eq:fused_pool} unchanged. Their weighted combination is therefore unchanged. The calibration denominator also stays unchanged, so the implementation's deterministic numerical convention gives the same result.

\end{document}